\documentclass[letterpaper, 10 pt, conference]{ieeeconf}  

\IEEEoverridecommandlockouts                              

\usepackage{graphics} 
\usepackage{epsfig} 
\usepackage{iftex}
\ifPDFTeX
  \usepackage{mathptmx}
\else
  \usepackage{fontspec}
\fi
\usepackage{amsmath} 
\usepackage{amssymb}  
\usepackage{cite}
\usepackage{url}
\usepackage{booktabs}
\usepackage{algorithm}
\usepackage{algorithmic}
\usepackage{multirow}
\usepackage{subfigure}
\usepackage{xcolor}

\newcommand{\task}[1]{\texttt{#1}}
\newcommand{\ours}{QDOS}

\title{\LARGE \bf
Learning Hierarchical Skill Policies with Offline Quality-Diversity Reinforcement Learning}

\author{Tanachai Anakewat$^{1}$, Takayuki Osa$^{2}$, and Tatsuya Harada$^{1,2}$\\
$^{1}$The University of Tokyo \quad $^{2}$RIKEN Center for Advanced Intelligence Project\\
Tokyo, Japan
}

\begin{document}

\maketitle
\thispagestyle{empty}
\pagestyle{empty}

\begin{abstract}

    Recent studies investigate how to leverage pre-collected datasets to improve the policy performance and sample efficiency of RL. One promising approach to achieve this goal is to employ a two-stage strategy: In the first stage, diverse skills are extracted as a low-level policy from a given dataset, and a high-level policy is trained to solve a specific task in the second stage. Typically, extraction of the low-level policy is performed based on unsupervised learning such as trajectory VAE. However, a limitation of this approach is that the quality of the low-level policy highly depends on the quality of the dataset. To address this issue, we introduce QDOS (Quality-Diversity Offline Skill learning), a unified pipeline for robust offline-to-online learning. Our approach incorporates an Advantage-Weighted Quality-Diversity pretraining objective, which weights the skill extraction and diversity objectives by the estimated advantage of each trajectory segment. This approach allows the model to extract diverse and high-value skills. By providing robust and task-relevant skill representations, QDOS significantly improves the quality of the embedded skill space used by the low-level policy. We further integrate this with a dual dataset reuse strategy, where offline data is used both for skill pretraining and for populating the online replay buffer via pseudo-labeling. Experiments demonstrate that QDOS significantly outperforms strong baselines in structured manipulation tasks and unstructured locomotion tasks, confirming its ability to accelerate exploration and improve final returns in challenging sparse-reward domains. 

\end{abstract}

\section{INTRODUCTION}

Deep reinforcement learning (RL) has achieved remarkable success in solving complex control tasks, ranging from playing video games to manipulating robotic arms \cite{mnih2015humanlevel,levine2016endtoend}. Despite these achievements, the standard online RL paradigm requires a large amount of interactions with the environment to learn effective policies. This sample inefficiency makes it prohibitively expensive and often unsafe for real-world applications, such as robotics, where data collection is costly and potentially dangerous.

To address these limitations, Offline RL has emerged as a promising alternative, enabling agents to learn policies entirely from static, pre-collected datasets without online interaction \cite{levine2020offline}. In parallel, unsupervised skill discovery methods aim to extract reusable behaviors, or ``skills,'' from large, unlabeled datasets. Recent methods such as OPAL~\cite{ajay2021opal} and SPiRL~\cite{pertsch2021skillpriors} encode primitive skills into a latent space and train a high-level policy that leverages these low-level policies to solve complex tasks. This temporal abstraction reduces the effective horizon of the task, simplifying the learning of long-horizon problems.

Recent work has begun to bridge these two fields in the ``offline-to-online" setting, where an agent is pretrained on offline data and then fine-tuned online interacting with the environment. Approaches such as SUPE \cite{wilcoxson2025supe} demonstrate that reusing offline data for both skill extraction and as a replay buffer for the high-level policy can dramatically accelerate exploration, resulting in higher performance after online RL.

\begin{figure}[t]
    \centering
    \includegraphics[width=0.9\linewidth]{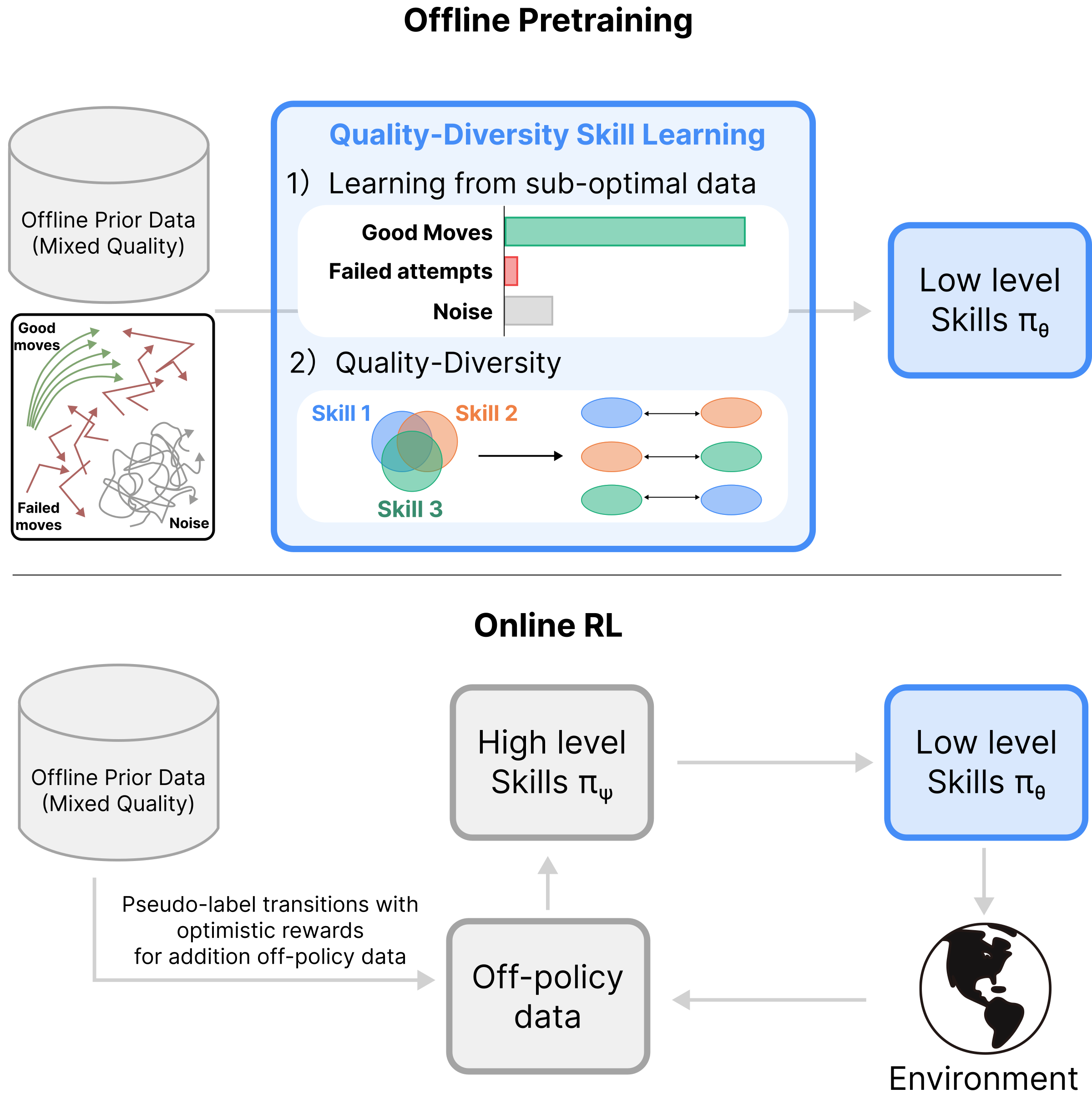}
    \caption{Overview of QDOS Framework. (1) \textbf{Offline Pretraining}: The agent learns low-level skills ($\pi_\theta$) from a mixed-quality dataset containing good moves, failed attempts, and noise. We employ an Advantage-Weighted Quality-Diversity objective to filter out sub-optimal data (failed attempts and noise) while extracting diverse, high-quality skills. (2) \textbf{Online RL with Dual Dataset Reuse}: The same offline data is reused to populate the high-level replay buffer via pseudo-labeling, enabling the high-level policy to utilize prior knowledge for efficient online exploration.}
    \label{fig:method_overview}
\end{figure}

However, a critical challenge remains: \textit{how to identify and extract useful behaviors when the offline data contains a mixture of optimal and suboptimal actions?} Real-world datasets often contain optimal trajectories entangled with noise, failed attempts, and irrelevant behaviors. Standard trajectory VAEs, as used in previous methods, treat all data segments equally, aiming to minimize the reconstruction error over the entire dataset. When the offline data is noisy, these methods learn a ``cluttered'' skill space where useful skills are entangled with useless ones, making it difficult for the high-level policy to select effective actions.

This leads to a trade-off in current approaches. Purely diversity-driven methods (e.g., maximizing mutual information) encourage distinct behaviors but fail to distinguish between distinct \textit{useful} behaviors and distinct \textit{useless} ones. Conversely, purely reward-driven or behavior-cloning approaches often collapse to a single mode, failing to capture the diversity of solutions necessary for robust downstream exploration. Consequently, there is a lack of a unified framework that can filter out noise while preserving a diverse set of high-quality skills suitable for accelerating online learning.

In this work, we propose \ours{} (Quality-Diversity Offline Skill learning), a unified framework for robust offline-to-online learning, as illustrated in Fig.~\ref{fig:method_overview}. Our method addresses the mixed-quality dilemma by introducing skill learning based on unsupervised multi-solution discovery using offline RL. The objective function for our skill learning incorporates both the expected return and a skill-diversity term, thereby yielding diverse and high-quality behaviors. This approach enables the learning of a skill space composed of useful primitives. We integrate this into an offline-to-online framework that reuses offline data twice: first for skill extraction, and second for populating the online replay buffer via ``pseudo-labeling," where offline trajectories are annotated with inferred skills and optimistic reward estimates following SUPE \cite{wilcoxson2025supe}. This allows the high-level policy to perform off-policy learning on the extensive offline dataset from the start.

We extensively evaluate our method on diverse and complex domains, including \task{antsoccer}, \task{kitchen}, and \task{humanoidmaze}. Our experiments demonstrate that our method significantly outperforms state-of-the-art baselines (such as SUPE \cite{wilcoxson2025supe}) in terms of sample efficiency, asymptotic performance, and goal-finding speed, particularly in tasks requiring the composition of diverse, dynamic behaviors.

\section{RELATED WORK}

\subsection{Unsupervised Skill Discovery}
Unsupervised skill discovery aims to learn reusable behaviors without extrinsic rewards, typically by maximizing an intrinsic objective. Early online methods focused on maximizing the mutual information between latent codes and states \cite{gregor2016variational,eysenbach2018diversity} or regularizing option policies \cite{bacon2017option}. These methods encourage the agent to visit diverse states and learn distinguishable behaviors.

In the offline setting, methods like OPAL \cite{ajay2021opal} and SPiRL \cite{pertsch2021skillpriors} adapt these ideas by using trajectory VAEs to learn skills from fixed demonstration datasets. They compress short trajectory segments into a latent variable $z$, allowing a high-level policy to operate in this latent space. However, these methods typically discard the offline data after the pretraining phase. Furthermore, they do not account for data quality; they try to reconstruct all trajectories, including suboptimal ones, which can degrade the quality of the learned skill space when the dataset is noisy.

\subsection{Offline Reinforcement Learning}
Offline RL focuses on learning optimal policies from static datasets. The primary challenge is distribution shift, where the learned policy visits states outside the dataset support, leading to overestimation of values. Algorithms like CQL \cite{kumar2020conservative} and IQL \cite{kostrikov2021offline} address this by constraining the policy or value function. While effective for learning from fixed data, these methods can be overly conservative, which hinders exploration when the agent is allowed to interact with the environment online.

\subsection{Offline-to-Online Reinforcement Learning}
Offline-to-online RL leverages offline data to accelerate online learning. A common strategy is to initialize an RL agent with a policy pretrained offline and then fine-tune it online \cite{ball2023efficient,lee2022offline}. Recent work like SUPE \cite{wilcoxson2025supe} proposes a more integrated approach, reusing offline data for both skill learning and online replay. By pseudo-labeling offline trajectories with inferred skills and rewards, SUPE allows the high-level policy to learn from the rich offline dataset. However, SUPE treats all offline data equally during skill learning, inheriting the brittleness of standard VAEs when dealing with mixed-quality data. Our method improves upon this by applying advantage weighting during skill discovery to ensure that only beneficial but diverse behaviors are captured and reused.

\subsection{Quality Diversity RL}
Quality-Diversity (QD) optimization is a paradigm that aims to generate large collections of diverse solutions that are all high-performing, contrasting with pure optimization which seeks a single global optimum. This concept was introduced by the Generative and Developmental Systems community~\cite{Lehman2011AbandoningOE,Mouret2015} with algorithms such as Novelty Search with Local Competition and MAP-Elites. QD algorithms operate in a behavioral or feature space rather than the genotypic space, attempting to fill the behavior space with high-performing solutions even if they are not global peaks in the fitness landscape.
In robotics, QD has been used to create repertoires of behaviors~\cite{Cully2015} and for damage adaptation. It has also been applied to engineering design~\cite{Gaier2018} and video game level generation~\cite{Khalifa2018Talakat}.

Recently, these ideas have been adapted to Reinforcement Learning. DiveOff~\cite{osa2024diveoff} introduces a QD perspective to offline RL, aiming to discover multiple solutions from a single task. DiveOff employs a variational objective that maximizes the mutual information between latent skills and trajectories to ensure diversity, while simultaneously regularizing the latent space to maintain high quality. Specifically, it uses a coordinate ascent approach to prevent mode collapse and ensure that the learned skills cover the manifold of useful behaviors. Our work builds upon this by explicitly integrating advantage-weighting into the QD objective, ensuring that diversity is sought specifically among high-value trajectories.

\subsection{Limitations of Current Approaches in Mixed-Quality Data}
Despite the promise of data reuse, existing methods like SUPE and OPAL rely on the assumption that the offline dataset consists of coherent, segmentable behaviors. In real-world scenarios, datasets are often unstructured, containing a mix of optimal trajectories, exploratory noise, and failed attempts. Standard trajectory VAEs minimize reconstruction error over the entire dataset, forcing the model to allocate representational capacity to noise. This results in a ``cluttered" skill space where distinct latent codes map to indistinguishable or useless behaviors. Our work addresses this critical gap by introducing an advantage-weighted objective that extracts high-quality segments and learns a latent skill space that corresponds to high-advantage behaviors.

\section{PRELIMINARIES}

\subsection{Problem Formulation}
We consider a Markov Decision Process (MDP) defined by the tuple $\mathcal{M} = \{\mathcal{S}, \mathcal{A}, \mathcal{P}, \gamma, r, \rho\}$, where $\mathcal{S}$ is the state space, $\mathcal{A}$ is the action space, $\mathcal{P}$ is the transition dynamics, $\gamma$ is the discount factor, $r$ is the reward function, and $\rho$ is the initial state distribution.

We assume access to a static offline dataset $\mathcal{D} = \{ \tau_i \}_{i=1}^{N}$, where each trajectory $\tau_i$ consists of a sequence of state-action-reward tuples $\tau_i = (s_0, a_0, r_0, \dots, s_T, a_T, r_T)$. Crucially, while rewards are present, the dataset is \textit{unlabeled} regarding skill definitions. There are no segmentations or labels $z$ indicating the active primitive behavior. The dataset contains both good and bad actions mixed together; the agent must infer which behaviors are worth learning and reusing.

The goal is two-fold:
\begin{enumerate}
    \item \textbf{Skill Discovery}: Learn a low-level latent-conditioned policy $\pi_\theta(a \mid s, z)$ and prior $p(z)$ that capture diverse, useful functional behaviors while filtering out noise and suboptimal data.
    \item \textbf{Hierarchical Adaptation}: Learn a high-level policy $\pi_\psi(z \mid s)$ that selects these skills to maximize expected return $\eta(\pi) = \mathbb{E}[\sum \gamma^t r_t]$ during online interaction.
\end{enumerate}

\subsection{Limitations of Current Approaches}
\begin{figure}[tb]
    \centering
    \includegraphics[width=0.9\linewidth]{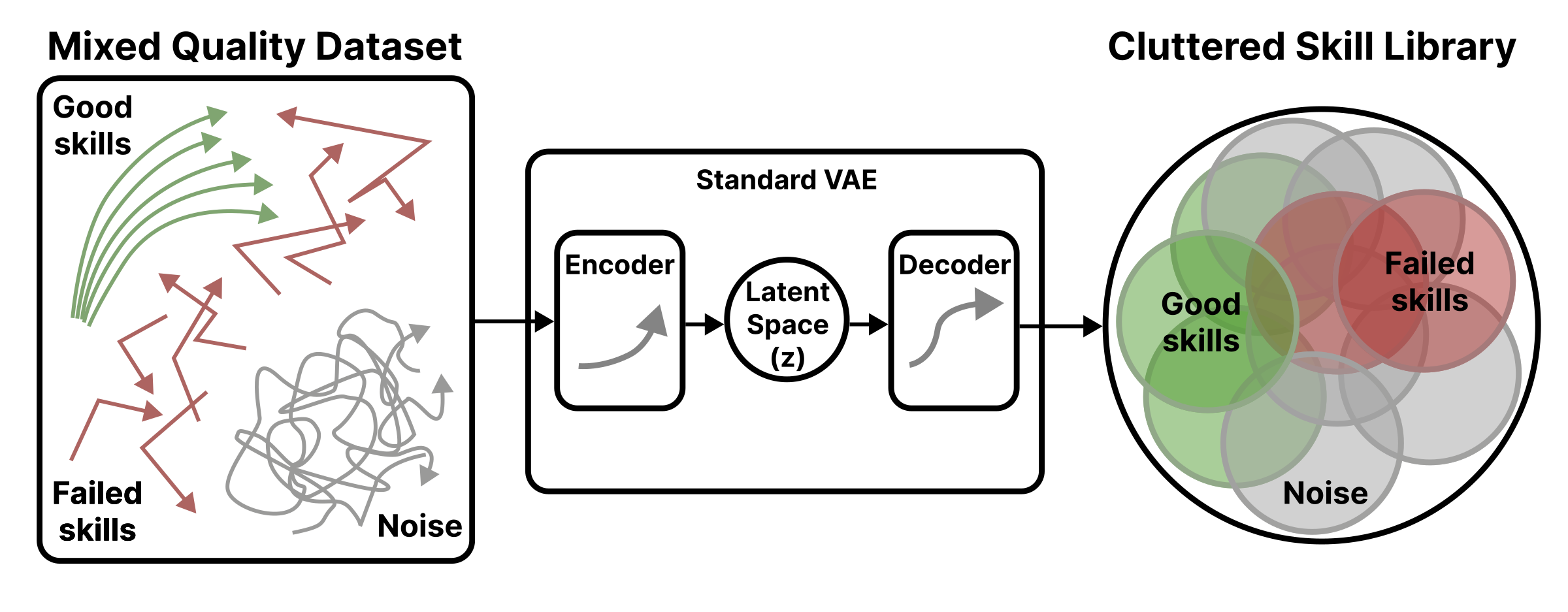}
    \caption{The Mixed-Quality Dilemma: Standard VAEs encode all behaviors equally, entangling noise with useful skills.}
    \label{fig:problem_statement}
\end{figure}

SUPE (Skills from Unlabeled Prior data for Exploration) \cite{li2024supe} uses a standard VAE to embed skills from the offline dataset and reuses this data to populate the online high-level replay buffer. While SUPE bridges offline skill learning and online RL, it implicitly assumes that the extracted skills are meaningful. Standard trajectory VAEs treat all segments as valid signals. In noisy datasets, the VAE reconstructs noise, resulting in a ``cluttered'' library where distinct latents map to indistinguishable or useless behaviors. This leads to unreliable pseudo-labeling, causing the high-level policy to receive inconsistent signals and fail to learn robust strategies. This mixed-quality issue is illustrated in Fig.~\ref{fig:problem_statement}.

\section{METHOD}

In this section, we detail our proposed framework, \ours{}, which consists of two main phases: (1) Advantage-Weighted Quality-Diversity Skill Pretraining, and (2) Online Exploration with Trajectory Skills. The overall procedure is summarized in Algorithm \ref{algo:qdos}.

\subsection{Advantage-Weighted Quality-Diversity Pretraining}
Our skill learning framework is inspired by DiveOff~\cite{osa2024diveoff}, which introduces a Quality-Diversity approach to offline RL. While the primary goal of DiveOff is to discover multiple diverse solutions within a static dataset, our objective is to \textit{reuse} these diverse, high-performing solutions to accelerate \textit{online} fine-tuning. By embedding these skills into a latent space, we provide the high-level policy with a structured action space that facilitates efficient exploration.

To effectively extract useful behaviors from mixed-quality datasets, we propose an Advantage-Weighted VAE framework. Standard VAEs minimize the reconstruction error over all data segments uniformly. This forces the model to dedicate capacity to reconstructing noise and suboptimal behaviors, cluttering the latent space. Instead, we propose to weight the learning objective by the estimated advantage of each trajectory segment.

\subsubsection{Advantage Estimation via IQL}
We first estimate the advantage of each trajectory segment. We segment the offline trajectories into fixed-length segments $\tau_{[H]} = \{s_0, a_0, \dots, s_{H-1}, a_{H-1}\}$ of length $H$. To estimate the quality of these segments, we train a value function $V_\psi(s)$ on the offline dataset using Implicit Q-Learning (IQL) \cite{kostrikov2021offline}. IQL learns a value function that approximates the upper expectile of the return distribution, effectively capturing the value of the best policy supported by the data without querying out-of-distribution actions.
The IQL value function is trained by minimizing the expectile regression loss:
\begin{equation}
L_{V}(\psi) = \mathbb{E}_{(s,a) \sim \mathcal{D}}[L_2^\tau(Q_{\hat{\psi}}(s,a) - V_\psi(s))],
\end{equation}
where $L_2^\tau(u) = |\tau - \mathbb{I}(u < 0)|u^2$ is the expectile loss with $\tau \in (0.5, 1)$ and $Q_{\hat{\psi}}$ denotes the target critic network. This allows us to estimate the potential return from any state in the dataset.
The advantage of a segment starting at state $s_0$ can then be estimated.
We compute a weight $w(\tau_{[H]})$ for each segment:
\begin{equation}
w(\tau_{[H]}) \propto \exp\left(\frac{A(\tau_{[H]})}{\lambda}\right),
\label{eq:advantage_weight}
\end{equation}
where $\lambda$ is a temperature parameter controlling the sharpness of the weight. High-advantage segments receive significantly higher weights, effectively focusing the VAE on learning from the ``good" parts of the dataset while ignoring the ``bad" parts. This filtering is crucial for preventing the skill space from being polluted by noise.

\subsubsection{Weighted VAE Objective}
We learn the skill space using a Variational Autoencoder (VAE). The encoder $f_\theta(z \mid \tau_{[H]})$ maps a trajectory segment to a latent variable $z$, and the decoder (low-level policy) $\pi_\theta(a \mid s, z)$ reconstructs the actions. We also learn a state-dependent prior $p_\theta(z \mid s_0)$ to capture the distribution of skills available at a given state. The weighted VAE objective is defined as:
\begin{align}
\mathcal{L}_{\text{VAE}}^{\text{AW}}(\theta) = \beta D_{\mathrm{KL}}(f_{\theta}(z \mid \tau_{[H]}) \parallel p_\theta(z \mid s_0)) \nonumber \\
- \mathbb{E}_{z \sim f_{\theta}}\left[ w(\tau_{[H]}) \sum_{t=0}^{H-1} \log \pi_{\theta}(a_t \mid s_t, z) \right].
\label{eq:weighted_vae}
\end{align}
This objective ensures that the model prioritizes the reconstruction of high-value behaviors. The KL-divergence term regularizes the learned posterior towards the prior, ensuring a smooth latent space.

\subsubsection{Mutual Information Maximization}
To prevent mode collapse (where the policy ignores the latent code) and ensuring diversity among the learned skills, we maximize the mutual information $I(z; \tau)$ between the latent skill $z$ and the generated trajectory $\tau$. Following variational information maximization principles \cite{osa2024diveoff, barber2004algorithm}, we utilize a variational lower bound:
\begin{equation}
I(z; \tau) \geq \mathbb{E}_{z \sim p(z), \tau \sim \pi_\theta(\cdot|z)} \left[ \log q_\theta(z \mid \tau) \right] + H(z),
\label{eq:mi_bound}
\end{equation}
where $q_{\theta}(z|\tau)$ is a parameterized posterior distribution.
Since the entropy $H(z)$ is constant for a fixed prior, maximizing mutual information is equivalent to maximizing the expected log-likelihood of the posterior. To maximize the entropy $H(z)$, a natural choice for the prior $p(z)$ is the uniform distribution. In practice, we sample random skills $z_{\text{rand}} \sim p(z)$, generate short auxiliary rollouts with the current policy $\pi_\theta$, and minimize the reconstruction error of the latent skill from these generated trajectories. Crucially, we also weight this diversity objective by the trajectory advantage. This ensures that we encourage diversity specifically among high-quality behaviors, rather than encouraging diverse failure modes. The expected log-likelihood of the posterior can be maximized by minimizing the following mean-squared L2 norm:
\begin{equation}
\mathcal{L}_{\text{info}}(\theta) = \frac{1}{B} \sum_{i=1}^B w^{(i)} \left\| \hat{z}_{\text{rand}}^{(i)} - z_{\text{rand}}^{(i)} \right\|_2^2,
\label{eq:info_loss}
\end{equation}
where $z_{\text{rand}}$ is sampled from the prior, and $\hat{z}_{\text{rand}}$ is encoded from the rollout of $\pi_\theta(z_{\text{rand}})$.

The total pretraining objective is a weighted combination:
\begin{equation}
\mathcal{L}_{\theta} = \mathcal{L}_{\text{VAE}}^{\text{AW}} + \alpha_{\text{loss}} \mathcal{L}_{\text{info}}.
\label{eq:total_loss}
\end{equation}

\subsection{Online Exploration with Trajectory Skills}
Following pretraining, the low-level skill policy $\pi_\theta$ is frozen. We then introduce a high-level policy $\pi_\psi(z \mid s)$ that operates in the latent skill space, selecting a new skill $z$ every $H$ environment steps.


To accelerate the learning of this high-level policy, we employ the dual dataset reuse strategy from SUPE \cite{wilcoxson2025supe}. We ``pseudo-label" the offline dataset $\mathcal{D}$ to create a replay buffer of high-level transitions. For each offline segment $\tau_{[H]}$, we:
1.  Infer the skill label $\hat{z} \sim q_\theta(z \mid \tau_{[H]})$ using the pretrained encoder.
2.  Compute an optimistic reward estimate $\hat{r}$ using an upper-confidence bound (UCB) estimator. We use Random Network Distillation (RND) \cite{burda2018exploration} to add an exploration bonus to the reward predicted by a learned reward model.

These pseudo-labeled transitions $(s_0, \hat{z}, \hat{r}, s_H)$ are added to the online replay buffer. This allows the high-level agent to perform off-policy learning (using an algorithm like RLPD \cite{ball2023efficient} or SAC \cite{haarnoja2018soft}) on the rich offline data from the very beginning of the online phase. The combination of high-quality, advantage-weighted skills and optimistic exploration bonuses guides the agent towards promising regions of the state space, significantly accelerating exploration.

\begin{algorithm}[t]
  \caption{\ours: Advantage-Weighted Quality-Diversity Skill Pretraining and Online RL}
  \label{algo:qdos}
  \scriptsize
  \linespread{1.04}\selectfont
  \setlength{\abovedisplayskip}{5pt}
  \setlength{\belowdisplayskip}{5pt}
  \setlength{\abovedisplayshortskip}{5pt}
  \setlength{\belowdisplayshortskip}{5pt}
  \begin{algorithmic}[1]
    \STATE \textbf{Inputs}: dataset $\mathcal{D}$, segment length $H$, batch size $B$, replay buffer $\mathcal{D}_{\mathrm{replay}} \leftarrow \emptyset$, temperature $\lambda$, mutual-information weight $\alpha$.
    \STATE \textbf{Phase 1: Advantage-Weighted Skill Pretraining}
    \FOR{each training step}
      \STATE Sample segments $\{\tau^{(i)}_{[H]}\}_{i=1}^{B} \sim \mathcal{D}$.
      \STATE Encode posterior $q_\theta(z \mid \tau^{(i)}_{[H]})$ and prior $p_\theta(z \mid s^{(i)}_0)$; sample $z^{(i)}$.
      \STATE Update IQL critics $Q_\psi, V_\psi$ to estimate values.
      \STATE Compute advantages $A_\psi(s^{(i)}_0, z^{(i)}) = Q_\psi(s^{(i)}_0, z^{(i)}) - V_\psi(s^{(i)}_0)$.
      \STATE Compute weights $w^{(i)} \propto \exp(A_\psi / \lambda)$ (normalized).
      \STATE \textbf{Weighted VAE loss}:
      \[
        \mathcal{L}_{\mathrm{VAE}}^{\mathrm{AW}} = \frac{1}{B} \sum_{i=1}^B w^{(i)} \Big[ \beta D_{\mathrm{KL}} - \mathbb{E}_{z}[\log \pi_\theta] \Big].
      \]
      \STATE \textbf{Mutual-information loss}: Sample $z_{\mathrm{rand}}^{(i)}$, rollout $\pi_\theta$, encode $\hat{z}_{\mathrm{rand}}^{(i)}$:
      \[
        \mathcal{L}_{\mathrm{info}} = \frac{1}{B} \sum_{i=1}^B w^{(i)} \left\| \hat{z}_{\mathrm{rand}}^{(i)} - z_{\mathrm{rand}}^{(i)} \right\|_2^2.
      \]
      \STATE Update $\theta$ via $\mathcal{L}_{\theta} = \mathcal{L}_{\mathrm{VAE}}^{\mathrm{AW}} + \alpha\,\mathcal{L}_{\mathrm{info}}$.
    \ENDFOR
    \STATE \textbf{Phase 2: Online RL with Pseudo-Labeling}
    \STATE \textbf{Initialize}: High-level policy $\pi_\psi(z|s)$ and $\mathcal{D}_{\mathrm{offline}} = \emptyset$.
    \FOR{each offline segment $\tau^{(i)}_{[H]} \in \mathcal{D}$}
        \STATE Infer $\hat{z}^{(i)} \sim q_\theta(z \mid \tau^{(i)}_{[H]})$ and optimistic reward $\hat{r}^{(i)}$.
        \STATE Add $(s^{(i)}_0, \hat{z}^{(i)}, \hat{r}^{(i)}, s^{(i)}_{H})$ to $\mathcal{D}_{\mathrm{offline}}$.
    \ENDFOR
    \FOR{each online interaction step}
      \STATE Sample $z \sim \pi_\psi(z \mid s)$, execute $\pi_\theta(a \mid s, z)$ for $H$ steps.
      \STATE Store transition in $\mathcal{D}_{\mathrm{replay}}$.
      \STATE Update $\pi_\psi$ using $\mathcal{D}_{\mathrm{replay}} \cup \mathcal{D}_{\mathrm{offline}}$ with off-policy RL (e.g., IQL).
    \ENDFOR
  \end{algorithmic}
\end{algorithm}

\subsection{Implementation Notes}
\label{sec:implementation}
We implement \ours{} using PyTorch. For skill pretraining, we use the Adam optimizer with a learning rate of $3 \times 10^{-4}$. The segment horizon is set to $H=20$ for all tasks. The advantage weights are computed using a value function trained via IQL on the offline dataset. The expectile parameter $\tau$ is fixed at $0.9$ for all environments, while the temperature parameter $\lambda$ is tuned per environment (typically $\lambda \in [0.3, 1.0]$). The mutual information weight $\alpha_{\text{loss}}$ is also a hyperparameter, with values in $\{0.1, 0.2, 0.3\}$ tested.

During online adaptation, we employ an entropy-regularized actor-critic (SAC) for the high-level policy. The replay buffer is initialized with the pseudo-labeled offline data, and we maintain a 50/50 sampling ratio between online and offline data during high-level updates to ensure stable learning. All experiments are averaged over 5 random seeds.

\section{EXPERIMENTS}

\subsection{Experimental Setup}
We evaluate \ours{} on four challenging sparse-reward domains that require long-horizon planning and precise control:
\begin{itemize}
    \item \textbf{\task{antmaze}}: A locomotion task where a quadrupedal ant must navigate a maze to reach a goal. We use the \texttt{large} maze layout.
    \item \textbf{\task{kitchen}}: A sequential manipulation task where a robotic arm must interact with multiple objects (e.g., microwave, kettle, switch) in a kitchen environment. We use the \texttt{mixed} and \texttt{partial} datasets.
    \item \textbf{\task{humanoidmaze}}: A high-dimensional locomotion task where a humanoid agent must navigate a maze. This task is significantly harder than AntMaze due to the stability requirements of the humanoid.
    \item \textbf{\task{antsoccer}}: A complex task where an ant must push a soccer ball to a goal location. This requires both locomotion and object manipulation skills.
\end{itemize}

We compare our method against several strong baselines:
\begin{itemize}
    \item \textbf{SUPE} \cite{wilcoxson2025supe}: Our primary baseline, a state-of-the-art offline-to-online method that uses trajectory skills without advantage weighting.
    \item \textbf{ExPLORe} \cite{li2024accelerating}: A method that uses exploration bonuses on offline data but learns a flat (non-hierarchical) policy.
    \item \textbf{Trajectory skills} \cite{ajay2021opal, pertsch2021skillpriors}: A baseline using standard trajectory-based skill learning without advantage weighting or diversity maximization.
    \item \textbf{IQL} \cite{kostrikov2021offline}: An offline RL method that learns a value function using expectile regression to avoid out-of-distribution actions.
    \item \textbf{BC}: Behavior Cloning, which learns a policy by maximizing the likelihood of actions in the offline dataset. For the learning curves, we compare against Diffusion BC with Jump-Start RL (JSRL) \cite{uchendu2023jump}, a strong baseline that initializes the policy with BC and fine-tunes it online.
\end{itemize}

We evaluate performance based on the normalized return (success rate for maze/soccer tasks, number of completed subtasks for Kitchen) and sample efficiency (number of steps to reach the goal).

\subsection{Overall Results}
\begin{figure}[tb]
    \centering
    \includegraphics[width=\linewidth]{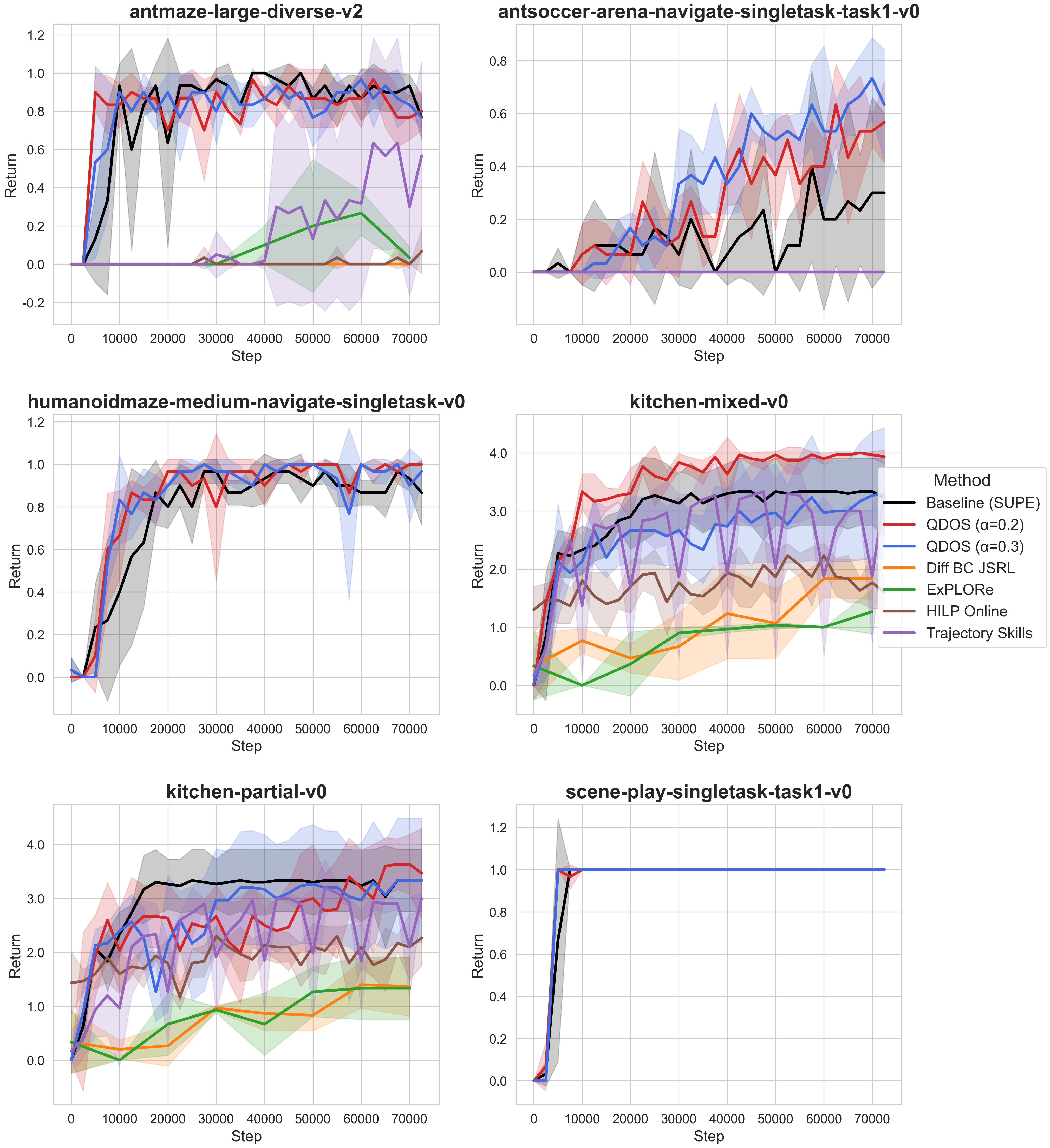}
    \caption{Learning curves across environments. QDOS is shown in red ($\alpha=0.2$) and blue ($\alpha=0.3$); Baseline (SUPE), ExPLORe, Trajectory Skills, HILP Online, and Diff BC JSRL are shown in black, green, purple, brown, and orange, respectively.}
    \label{fig:learning_curves}
\end{figure}

\ours{} consistently outperforms baselines across all evaluated domains. The quantitative results are summarized in Table \ref{tab:overall-results}, and the training dynamics are shown in Fig.~\ref{fig:learning_curves}.

\begin{table*}[t]
    \centering
    \caption{Normalized evaluation returns comparing BC, IQL, SUPE, and \ours{} across various domains. Results for SUPE and \ours{} are averaged over 3 seeds.}
    \label{tab:overall-results}
    \resizebox{0.8\textwidth}{!}{%
    \begin{tabular}{lccccc}
        \toprule
        \textbf{Environment} & \textbf{BC}\cite{kostrikov2021offline}\cite{ogbench_park2024} & \textbf{IQL}\cite{kostrikov2021offline}\cite{ogbench_park2024}  & \textbf{SUPE} & \textbf{\ours{} ($\alpha=0.2$)} & \textbf{\ours{} ($\alpha=0.3$)} \\
        \midrule
        \task{antmaze-large} & $41 \pm 7$ & $64 \pm 10 $ & $83.0 \pm 21.0$ & $\underline{86.0 \pm 6.0}$ & $\mathbf{96.0 \pm  6.0}$\\
        \task{antsoccer-arena} & $5 \pm 1$ & $50\pm2$ & $27.0 \pm 31.0$ & $\underline{73.0 \pm 31.0}$ & $\mathbf{80.0 \pm 10.0}$ \\
        \task{humanoidmaze-medium} & $8\pm2$ & $27\pm2$ & $\underline{97.0 \pm 6.0}$ & $\underline{97.0 \pm 6.0}$ & $\mathbf{100.0 \pm 0.0}$ \\
        \task{kitchen-mixed} & 51.5 & 51.0 & $\underline{85 \pm 13.25}$ & $\mathbf{100.00 \pm 0.00}$ & $83.25 \pm 28.75$ \\
        \task{kitchen-partial} & 38.0 & 46.3 & $\mathbf{83.25 \pm 14.50}$ & $\mathbf{85.75 \pm 24.75}$ & $\mathbf{82.50 \pm 30.25}$ \\
        \task{scene-task1} & $5\pm1$ & $51\pm4$ & $\mathbf{100.0 \pm 0.0}$ & $\mathbf{100.0 \pm 0.0}$ & $\mathbf{100.0 \pm 0.0}$ \\
        \bottomrule
    \end{tabular}%
    }
\end{table*}

In the highly complex \task{antsoccer-arena} environment, \ours{} achieves a normalized return of $\mathbf{0.80}$, which is a substantial improvement over SUPE's $0.27$. This task requires fine-grained control to manipulate the ball, and standard skill discovery methods likely fail to capture these precise interactions amidst the noise of general locomotion. Our advantage-weighting ensures that the skills focus on the moments where the ball is successfully moved.

In \task{kitchen-mixed}, \ours{} attains a perfect score of $\mathbf{4.00}$, effectively stitching together fragmented subtasks from the dataset to solve the full sequential task. The baseline SUPE achieves only 3.40, indicating it struggles to compose the necessary skills or fails to learn some subtasks entirely.

In \task{humanoidmaze}, our method achieves perfect performance ($\mathbf{1.00}$), slightly outperforming SUPE. This confirms that even in high-dimensional state spaces, filtering for high-quality skills leads to more robust locomotion.

\subsection{Goal Finding Efficiency}
A key advantage of our method is the speed at which it discovers solutions. In sparse-reward tasks, the time to reach the first goal is a critical metric for exploration efficiency. We analyzed the number of environment steps required to reach the goal for the first time.

\begin{figure}[tb]
    \centering
    \includegraphics[width=0.9\linewidth]{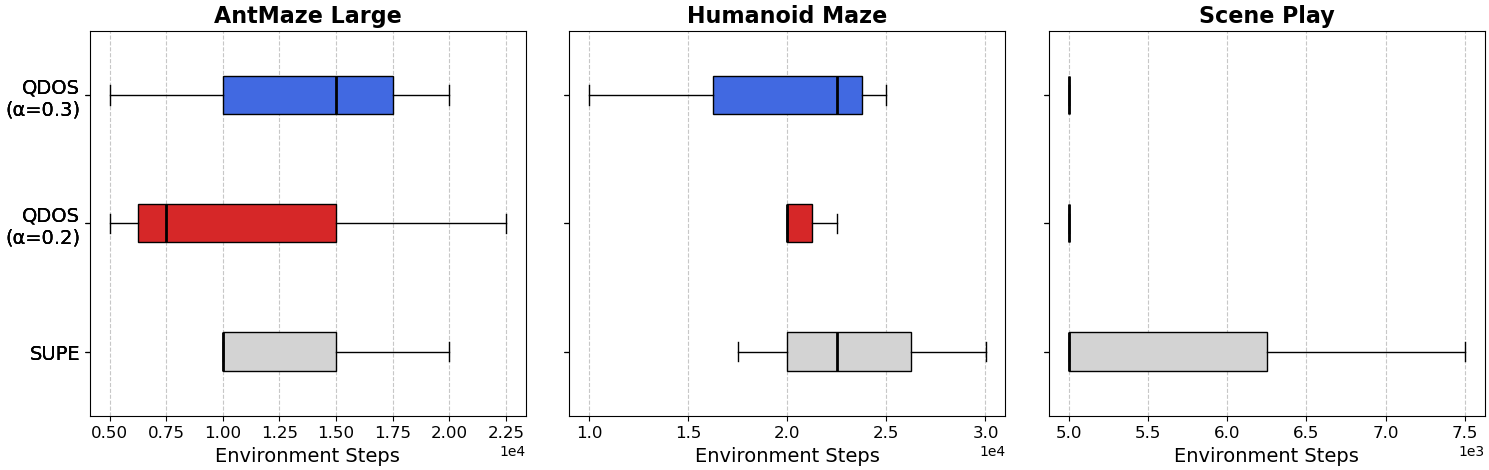}
    \caption{Training steps to first goal finding. QDOS (Red) finds the goal significantly faster than baselines, indicating more efficient exploration.}
    \label{fig:steps_to_goal}
\end{figure}

\begin{table}[h]
    \centering
    \caption{Training steps to reach the goal for the first time (Mean $\pm$ Std).}
    \label{tab:steps_to_goal}
    \begin{tabular}{l|l|c}
        \toprule
        \textbf{Environment} & \textbf{Method} & \textbf{Steps} \\
        \midrule
        \multirow{3}{*}{\task{antmaze-large}} & SUPE & $13334 \pm 5774$ \\
         & \ours{} ($\alpha=0.2$) & $\mathbf{11668 \pm 9465}$ \\
        \hline
        \multirow{3}{*}{\task{humanoidmaze}} & SUPE & $23334 \pm 6292$ \\
         & \ours{} ($\alpha=0.3$) & $\mathbf{19168 \pm 8036}$ \\
        \hline
        \multirow{3}{*}{\task{scene}} & SUPE & $5834 \pm 1443$ \\
         & \ours{} ($\alpha=0.2$) & $\mathbf{5001 \pm 0}$ \\
        \bottomrule
    \end{tabular}
\end{table}

As shown in Table \ref{tab:steps_to_goal} and Fig.~\ref{fig:steps_to_goal}, \ours{} consistently finds the goal faster than SUPE. For instance, in the \task{scene} task, \ours{} reaches the goal in the minimum possible time ($5001$ steps), while SUPE takes significantly longer ($5834$ steps). In \task{humanoidmaze}, our method reduces the exploration time by over 4000 steps. This accelerated exploration is a direct result of the cleaner skill space: the high-level policy does not waste time exploring with noisy or ineffective skills, allowing it to traverse the environment more efficiently.

\subsection{Hyperparameter Sensitivity}
We analyzed the sensitivity of our method to the mutual information weight $\alpha_{\text{loss}}$, which balances the reconstruction quality and the diversity of the skills. We observed a task-dependent trade-off.

For complex, dynamic tasks like \task{antsoccer}, a higher diversity weight ($\alpha_{\text{loss}}=0.3$) yields the best performance. The solution space for manipulating a soccer ball is large, and encouraging a diverse range of interaction skills increases the likelihood of finding a successful strategy.

Conversely, for structured manipulation tasks like \task{kitchen}, a moderate diversity weight ($\alpha_{\text{loss}}=0.2$) performs better. In these tasks, the optimal behaviors are more constrained (e.g., opening a microwave requires a specific motion). Excessive diversity can lead to the learning of task-irrelevant variations that distract the policy. This suggests that \ours{} is flexible and can be tuned to the specific needs of the domain.

\subsection{Failure Case}
\begin{figure}[tb]
    \centering
    \includegraphics[width=0.9\linewidth]{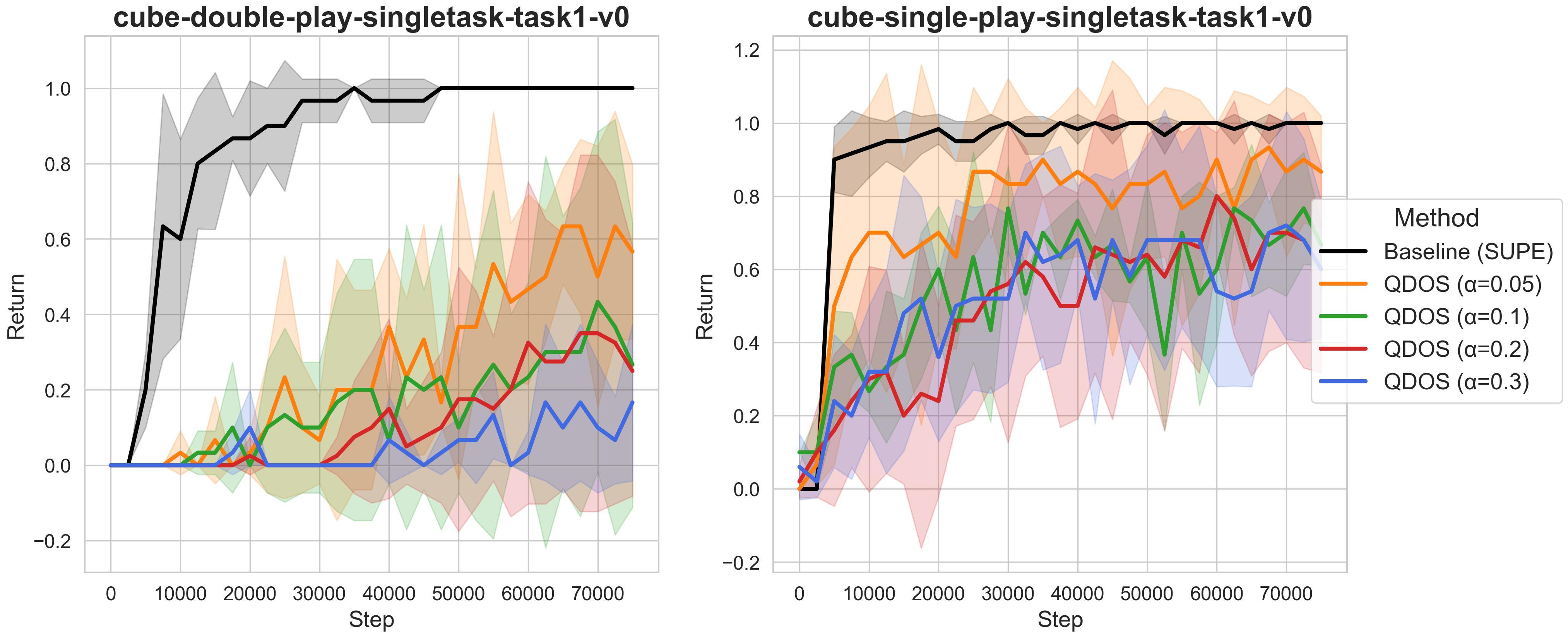}
    \caption{Experimental results on the \task{cube} datasets. Reducing the diversity weight $\alpha_{\text{loss}}$ leads to performance closer to the SUPE baseline, indicating that explicit diversity enforcement is less critical for high-quality datasets.}
    \label{fig:cube_graph}
\end{figure}

We extended our hyperparameter analysis to high-quality datasets such as \task{cube-single} and \task{cube-double} that only require pick-and-place motion. As illustrated in Fig. ~\ref{fig:cube_graph}, we observed that reducing the mutual information weight $\alpha_{\text{loss}}$ causes our method to converge towards the performance of the baseline. This trend reinforces our finding that for datasets with high-quality, structured demonstrations, the need for enforcing diversity via mutual information is diminished.

\section{VISUALIZATION OF THE MODEL}

To better understand the properties of the learned skill space and how they contribute to performance, we conducted qualitative visualizations of the latent space and the resulting agent behaviors.

\subsection{Latent Space Structure}
\begin{figure*}[tb]
    \centering
    \includegraphics[width=0.8\linewidth]{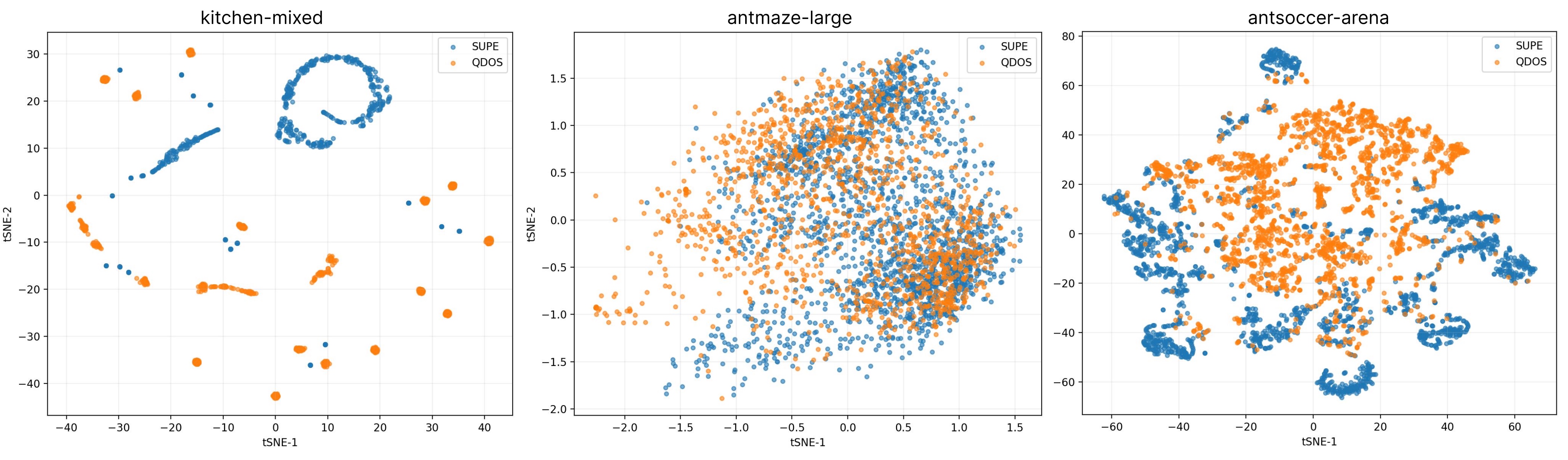}
    \caption{t-SNE visualization (top) and skill usage timeline (bottom) across \task{kitchen}, \task{antmaze}, and \task{antsoccer}. Orange points represent \ours{} ($\alpha=0.3$ for \task{antsoccer} and \task{antmaze}, $\alpha=0.2$ for \task{kitchen}), and Blue points represent SUPE. The t-SNE plots indicate that \ours{} tends to utilize a broader region of the latent space, suggesting the capture of more diverse behaviors compared to the more restricted latent usage of SUPE.}
    \label{fig:skill_visualization}
\end{figure*}

We analyzed the structure of the learned latent space by projecting the latent variables from the offline dataset into a 2D space using t-SNE, as shown in Fig.~\ref{fig:skill_visualization}. The visualization compares the latent space usage of \ours{} (Orange) and SUPE (Blue) across \task{kitchen}, \task{antmaze}, and \task{antsoccer} tasks. We observe that the latent variables for \ours{} appear to be distributed more broadly across the space compared to SUPE. This trend suggests that \ours{} might be utilizing the latent space more extensively to represent a diverse set of skills, whereas SUPE seems to use the latent space in a more limited or concentrated manner.

\subsection{Behavior Visualization}
\begin{figure}[t!]
    \centering
    \includegraphics[width=0.9\linewidth]{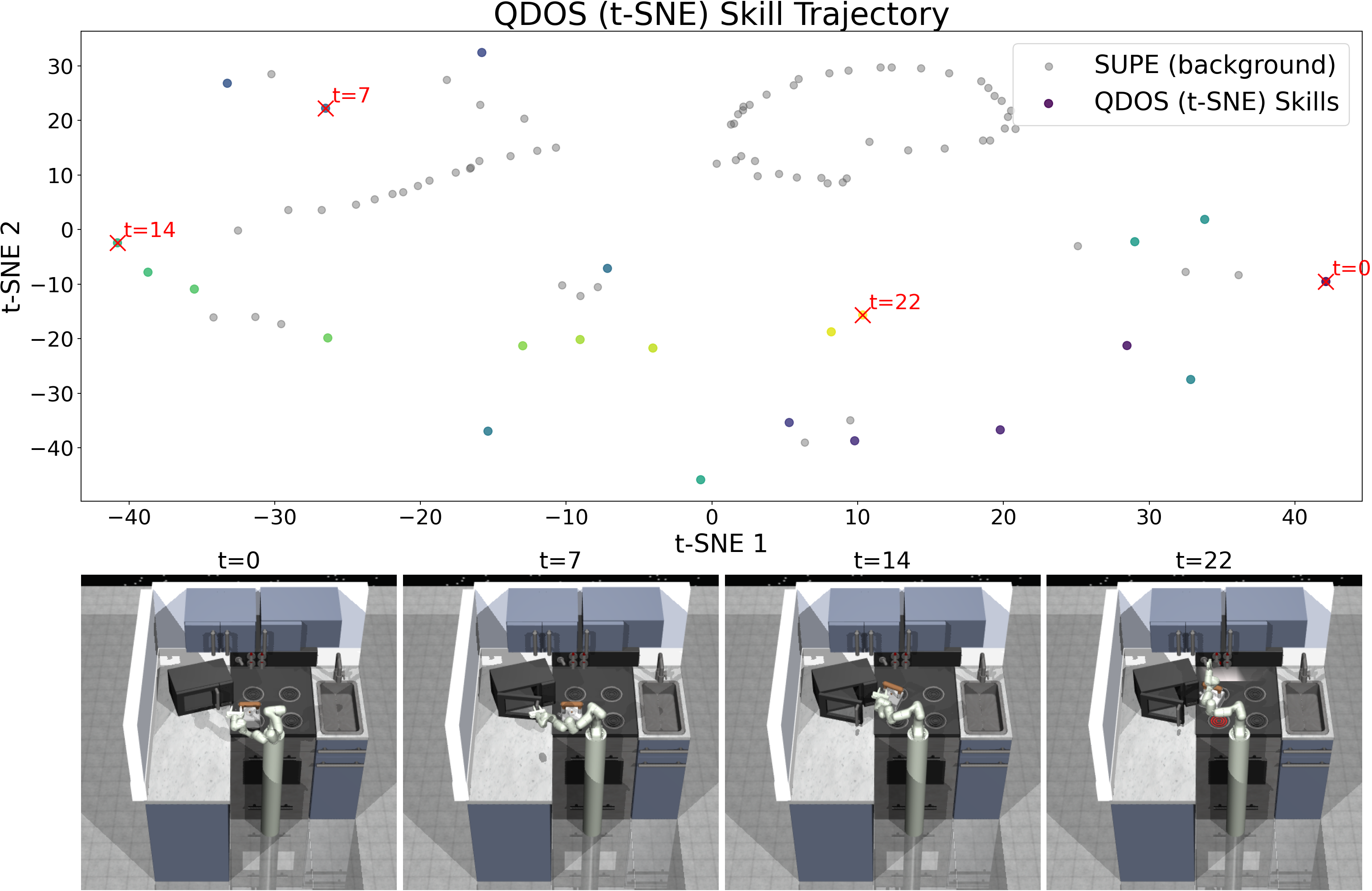}
    \caption{QDOS skill trajectory visualization for \task{kitchen-mixed}. Red crosses mark selected timesteps ($t=0, 7, 14, 22$) with corresponding scene snapshots.}
    \label{fig:qdos_tsne}
\end{figure}

We also visualized the rollout trajectories of the learned skills in the \task{kitchen} and \task{antmaze} domains. In \task{kitchen}, the learned skills correspond to distinct, semantically meaningful object interactions. For example, specific latent codes consistently trigger the ``open microwave" or ``move kettle" behaviors. The pseudo-labeling process accurately assigns these skills to the corresponding offline segments, enabling the high-level policy to compose them effectively. As shown in Fig.~\ref{fig:qdos_tsne}, QDOS reaches the goal at $t=22$ in the \task{kitchen-mixed} trajectory, and the trajectory pattern in latent space is clearly structured, which is consistent with effective skill composition.

In \task{antmaze}, the skills represent coherent directional movements (e.g., ``move north", ``turn left"). By continuously manipulating the continuous latent variable $z$ (e.g., interpolating between two latent codes), we observed smooth transitions between behaviors, such as gradually changing the movement direction or speed. This confirms that the latent space captures the underlying manifold of useful motions, providing a smooth and navigable action space for the high-level policy. This smoothness is crucial for the stability of the high-level RL training.

\section{CONCLUSION}

We presented \ours{}, a unified framework for Quality-Diversity offline-to-online Reinforcement Learning. By integrating advantage weighting into the skill pretraining objective, \ours{} effectively addresses the challenge of learning from mixed-quality offline datasets. It distills robust, task-aligned skills while filtering out noise and failures, overcoming the limitations of standard unsupervised skill discovery methods.

We demonstrated that combining this robust skill learning with a dual dataset reuse strategy, using offline data for both skill extraction and high-level initialization, leads to significant performance gains. Empirical results on challenging domains like \task{antsoccer}, \task{kitchen}, and \task{humanoidmaze} show that \ours{} outperforms state-of-the-art baselines in terms of asymptotic return, sample efficiency, and exploration speed.

These findings highlight the importance of quality-aware representation learning in offline RL. Future work will explore adaptive methods to automatically tune the quality-diversity trade-off and investigate mechanisms for fine-tuning the low-level skills during online interaction to further adapt to novel situations.

\appendix

\subsection{Environment Details}
We evaluate our method on three distinct domains, each presenting unique challenges for skill discovery and hierarchical control.

\subsubsection{State-based Locomotion}
\task{antmaze}, \task{humanoidmaze}, and \task{antsoccer} involve controlling complex agents to navigate mazes or manipulate objects.
\begin{itemize}
    \item \task{antmaze}: A standard benchmark from D4RL~\cite{fu2020d4rl} where a quadrupedal ant must navigate a large maze. We use the \texttt{large} maze layout, which requires long-horizon planning.
    \item \task{humanoidmaze}: A domain from OGBench~\cite{ogbench_park2024} featuring a high-dimensional humanoid agent. The agent must balance while navigating, making the skill space significantly more complex than AntMaze.
    \item \task{antsoccer}: A multi-stage task where an ant must navigate to a soccer ball and push it to a goal. This requires composing locomotion skills with object interaction dynamics.
\end{itemize}

\subsubsection{State-based Manipulation}
\task{kitchen}, \task{cube}, and \task{scene} focus on robotic manipulation.
\begin{itemize}
    \item \task{kitchen}: A sequential manipulation task from D4RL~\cite{fu2020d4rl} where a 7-DOF Franka robot must manipulate multiple objects (microwave, kettle, burner, switch) in sequence. We use the \texttt{mixed} dataset, which contains various subtasks being performed, but the 4 target subtasks are never completed in sequence together. We also use the \texttt{partial} dataset, which includes other tasks being performed, but there are sub-trajectories where the 4 target subtasks are completed in sequence.
    \item \task{cube}: A manipulation domain from OGBench~\cite{ogbench_park2024} focusing on pick-and-place tasks. We use \task{cube-single} which requires lifting a single cube, and \task{cube-double} which involves manipulating two cubes.
    \item \task{scene}: A domain from OGBench~\cite{ogbench_park2024} requiring the composition of atomic behaviors like locking/unlocking and opening drawers.
\end{itemize}

\subsection{Hyperparameters}
We provide the detailed hyperparameters used for the VAE skill pretraining and the high-level online RL agent in Table~\ref{tab:vae-hyperparams} and Table~\ref{tab:rl-hyperparams}, respectively.

\begin{table}[h]
    \centering
    \begin{minipage}{0.7\columnwidth}
        \centering
        \caption{Hyperparameters for VAE training.}
        \label{tab:vae-hyperparams}
        \resizebox{\columnwidth}{!}{%
        \begin{tabular}{l|c}
            \toprule
            \textbf{Parameter Name} & \textbf{Value} \\
            \midrule
            Batch size & $256$ \\
            Optimizer & Adam \\
            Learning rate & $3 \times 10^{-4}$ \\
            GRU Hidden Size & 256 (\task{antmaze}, \task{kitchen}) \\
            & $512$ (\task{antsoccer}, \task{humanoidmaze}, \task{scene}) \\
            GRU Layers & $2$ (\task{antmaze}, \task{kitchen}) \\
            & $3$ (\task{antsoccer}, \task{humanoidmaze}, \task{scene}) \\
            KL Coefficient ($\beta)$ & $0.1$ (\task{antmaze}, \task{humanoid}, \task{kitchen}) \\
            & $0.2$ (\task{scene}) \\
            Latent Dimension ($z$) & $8$ \\
            Segment Length ($H$) & $20$ \\
            \bottomrule
        \end{tabular}
        }
    \end{minipage}
    
    \vspace{0.5cm}

    \begin{minipage}{0.7\columnwidth}
        \centering
        \caption{Hyperparameters for the high-level online RL agent.}
        \label{tab:rl-hyperparams}
        \resizebox{\columnwidth}{!}{%
        \begin{tabular}{l|c}
            \toprule
            \textbf{Parameter} & \textbf{Value} \\
            \midrule
            Batch size &  256 \\
            Discount factor ($\gamma$) & 0.99 (AntMaze, Kitchen) \\
            & 0.995 (Humanoid, Scene, Soccer) \\
            Optimizer & Adam \\
            Learning rate & $3 \times 10^{-4}$ \\
            Critic ensemble size & 10 \\
            UTD Ratio & 20 \\
            Network Width & 256 (AntMaze, Kitchen) \\
            & 512 (Humanoid, Soccer, Scene) \\
            Network Depth & 3 hidden layers \\
            RND coefficient ($\alpha$) & $8.0$ \\
            \bottomrule
        \end{tabular}
        }
    \end{minipage}
\end{table}

\subsection{Acknowledgement}
This work was partially supported by JST Moonshot R\&D Grant Number JPMJPS2011, CREST Grant Number JPMJCR2015 and Basic Research Grant (Super AI) of Institute for AI and Beyond of the University of Tokyo.
Takayuki Osa was supported by JSPS KAKENHI Grant Number JP25K03176.

\subsection{Use of Large Language Models (LLMs)}
We utilized large language models to refine the clarity and grammar of the text throughout this manuscript. All scientific content, data, and figures are the original work of the authors.

\bibliographystyle{IEEEtran}
\begingroup
\renewcommand{\baselinestretch}{0.98}\selectfont
\bibliography{IEEEabrv,mybibfile}
\endgroup

\end{document}